\documentclass[a4paper]{article}

\usepackage{INTERSPEECH2016}

\usepackage{graphicx}
\usepackage{amssymb,amsmath,bm}
\usepackage{textcomp}
\usepackage{color}

\ninept

\title{\vspace*{-4pt}Training variance and performance evaluation of neural networks in speech}

\makeatletter
\def\name#1{\gdef\@name{#1\\}}
\makeatother \name{{\em Ewout van den Berg$^*$, Bhuvana
    Ramabhadran$^*$, Michael Picheny$^*$}}

\address{$^*$IBM Watson Group \\
  {\small \tt evanden@us.ibm.com, bhuvana@us.ibm.com, picheny@us.ibm.com}
}

\begin{document}

  \maketitle
  \begin{abstract}
In this work we study variance in the results of neural network
training on a wide variety of configurations in automatic speech
recognition. Although this variance itself is well known, this is, to
the best of our knowledge, the first paper that performs an extensive
empirical study on its effects in speech recognition. We view training
as sampling from a distribution and show that these distributions can
have a substantial variance. These results show the urgent need to
rethink the way in which results in the literature are reported and
interpreted\footnote{Part of this work was presented at ICLR 2016.}.
  \end{abstract}

\section{Introduction}
In automatic speech recognition (ASR), the goal is to develop a
combination of language and acoustic models that together minimize the
decoding word-error rate (WER) on predefined tasks. Early work on the
application of deep learning for ASR showed tremendous improvement of
up to 33\% relative to existing GMM-HMM
models~\cite{SEI2011LYa}. Results like these led to a proliferation of
research on variations of deep neural network architectures along with
new training methods, alternative feature representations and ways to
improve data augmentation.  The performance of newly proposed
approaches is typically evaluated by comparing the results against the
performance of a baseline system. Even after extensive tuning of the
hyperparameters of the new system, relative improvements of around 5\%
are much more common now. In this work we look at the well-known but
widely-ignored issue of variance in neural network training and its
implications on evaluating new methods. In Section~\ref{Sec:Sampling}
we look at neural network training as sampling from a distribution and
in Section~\ref{Sec:NumericalExperiments} we perform extensive
experiments that empirically show what these distributions look like
for practical ASR tasks. In Section~\ref{Sec:Discussion} we discuss
the implications on the evaluation of new methodologies, and conclude
in Section~\ref{Sec:Conclusions}.

\section{Neural network training as sampling}\label{Sec:Sampling}

For the vast majority of conventional neural networks, training
involves minimizing a cost function that is highly non-convex. Given
that optimization is done with techniques that were designed for
convex problems, such as stochastic gradient descent (SGD), it should
come as no surprise that the solution of a training run depends on the
initial starting point, as well as various factors that affect the
optimization process, such as the mini-batch randomization (see also
\cite{CHO2014HMAa,PAS2014DGBa,PIN2009DDCa}). This phenomenon is common
knowledge to many practitioners, and indeed it was leveraged in work
by \cite{HIN2015VDa} to generate model ensembles. However, it is
otherwise rarely discussed explicitly in the literature

Training algorithms are generally deterministic given the initial
state of the network and the order of the training data (we exclude
asynchronous algorithms, in which the exact timing and communication
network load may affect the outcome). The initial state and order in
turn are often deterministic given the state $s$ of the pseudo-random
number generator (PRNG). Denoting by $\theta$ the model setup,
including the network architecture, network parameters, training
algorithm, and hyper parameters, we can interpret training as
evaluating the function $M(\theta, s)$, mapping the training setup and
PRNG state to a trained model. When the state of the random number
generator is not explicitly controlled, $s$ can be viewed as a hidden
random variable $s \sim \mathcal{S}$. As a consequence, $M(\theta,s)$
is turned into a distribution $\mathcal{M}(\theta)$. Training of a
network then changes from evaluating a function to drawing a sample
$m\sim\mathcal{M}(\theta)$\footnote{Strictly speaking, $s$ is sampled
  from $\mathcal{S}$ only when the program is started, and assuming a
  suitable initialization based on time and random sources. When
  training two networks in a row the successive states will not be
  independent.}

\section{Numerical experiments}\label{Sec:NumericalExperiments}

\subsection{Experiment setup}
For the training and evaluation of the systems we use three datasets.
The first two datasets are based on the English Broadcast News (BN)
training corpus \cite{FIS1997GPFa} which we pre-process to obtain
40-dimensional logmel features with speaker-dependent mel filters
chosen from a set of 21 possible filters. The first dataset (BN400)
contains the complete 400-hour BN training set and a 30-hour hold-out
set. The second dataset (BN50) uses only a subset of the data and
defines a 45-hour training set and a 5-hour hold out set
\cite{KIN2009a}. For evaluation we use EARS Dev-04f, as described by
\cite{KIN2009a}. The third dataset is based on the 300h Switchboard
corpus \cite{GOD1997Ha,LIU1995MNPa} and uses 40-dimensional
speaker-independent logmel features. Evaluation is done on the
Hub5-2000 and CallHome corpora \cite{SAO2015KRPa}. For all decodes we
use the trigram language model (LM) as described in \cite{KIN2009a}.

For the acoustic model we consider both DNN and CNN architectures,
each with a softmax output layer mapping to 5999 tri-phone states for
BN and 9300 tri-phone states for Switchboard. The DNN has an input
layer that takes the logmel features with $\pm 4$ temporal context for
BN and $\pm 5$ for SWB. This input is then transformed through five
hidden linear layers, each with a sigmoid activation function and an
output dimension of 1024 for BN and 2048 for SWB, before reaching a
softmax output layer. The CNN consists of two convolution layers (with
max pooling and respectively 128 filters of size $9\!\times\!
9\!\times\!  3$ and 25 of size $3\!\times\!  4\!\times\! 128$), two
hidden linear layers with sigmoid activation of size 1024, and a final
output layer. The input features are logmel with $\pm 4$ frames
context and $\Delta$ and $\Delta^2$ information.

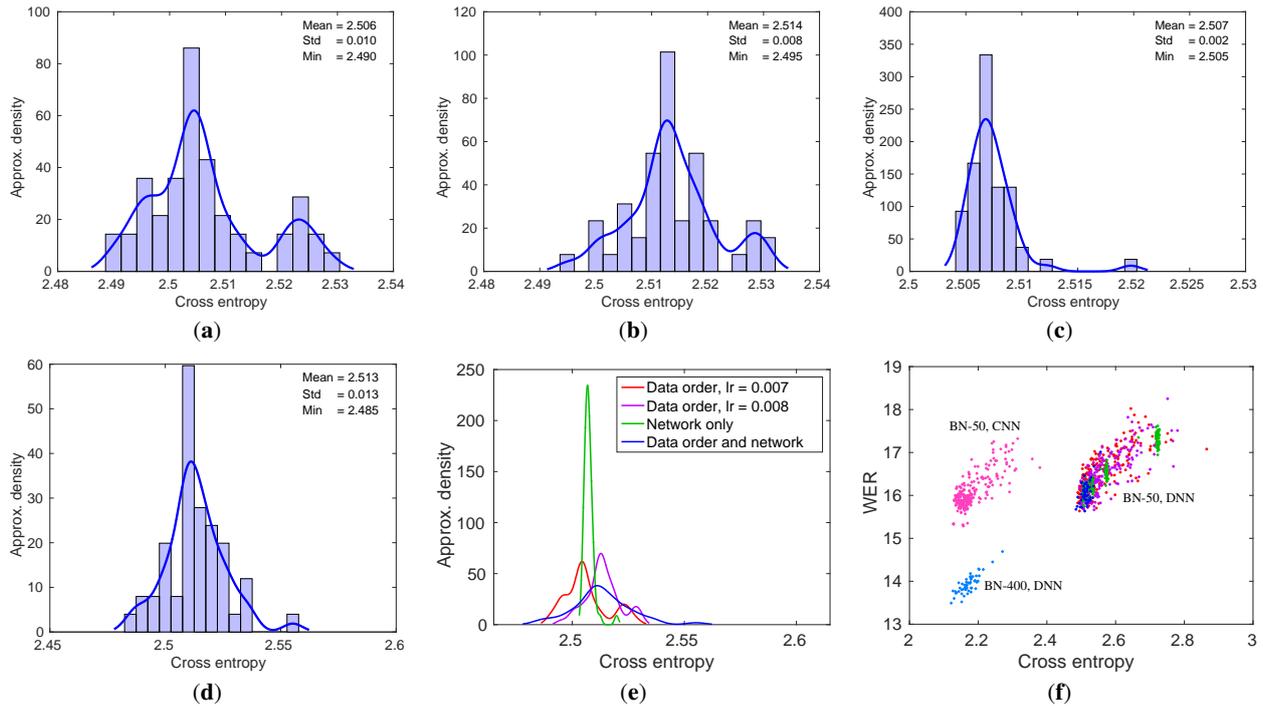
\begin{figure*}[t]
\centering
\begin{tabular}{ccc}
\includegraphics[width=0.305\textwidth]{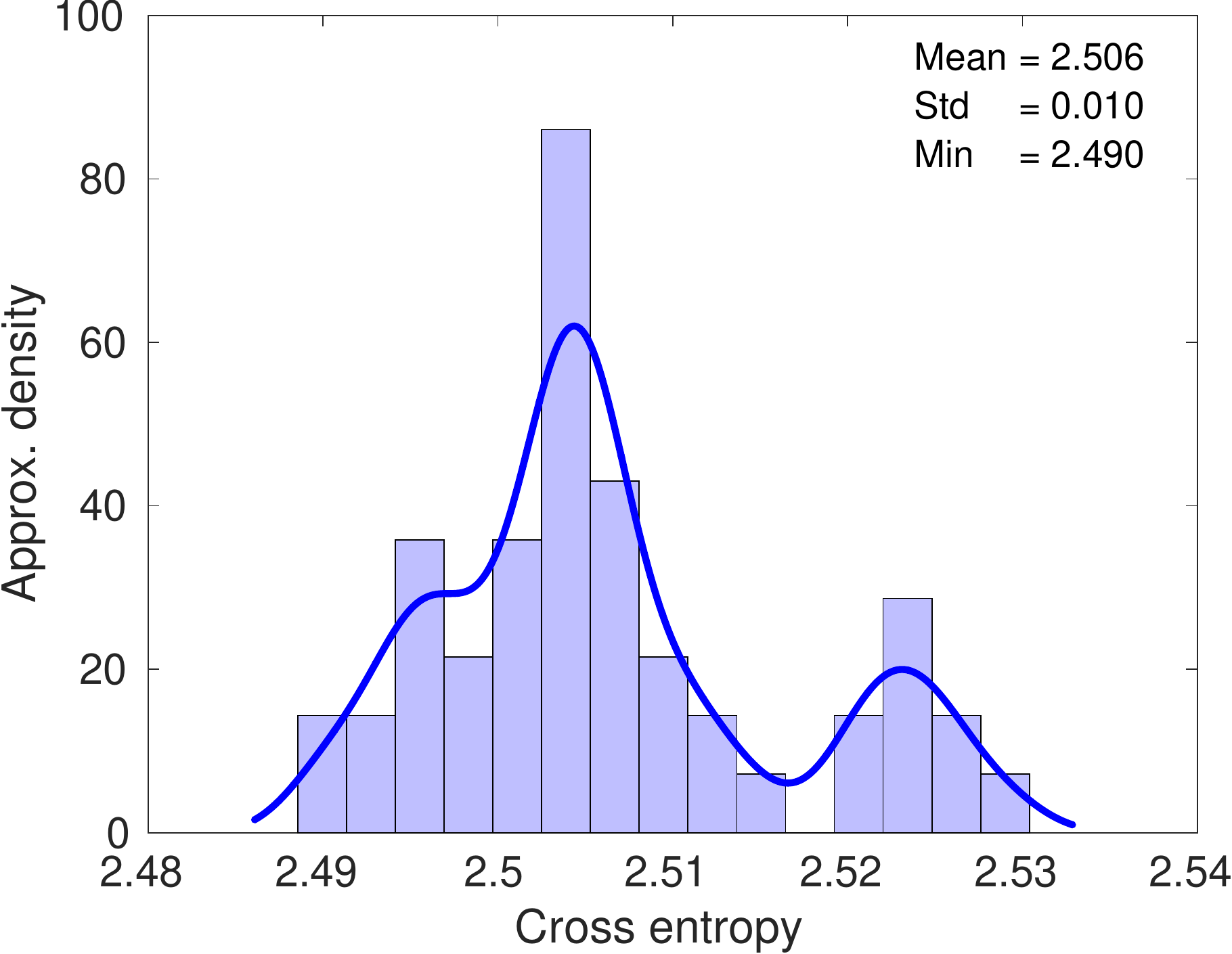}&
\includegraphics[width=0.305\textwidth]{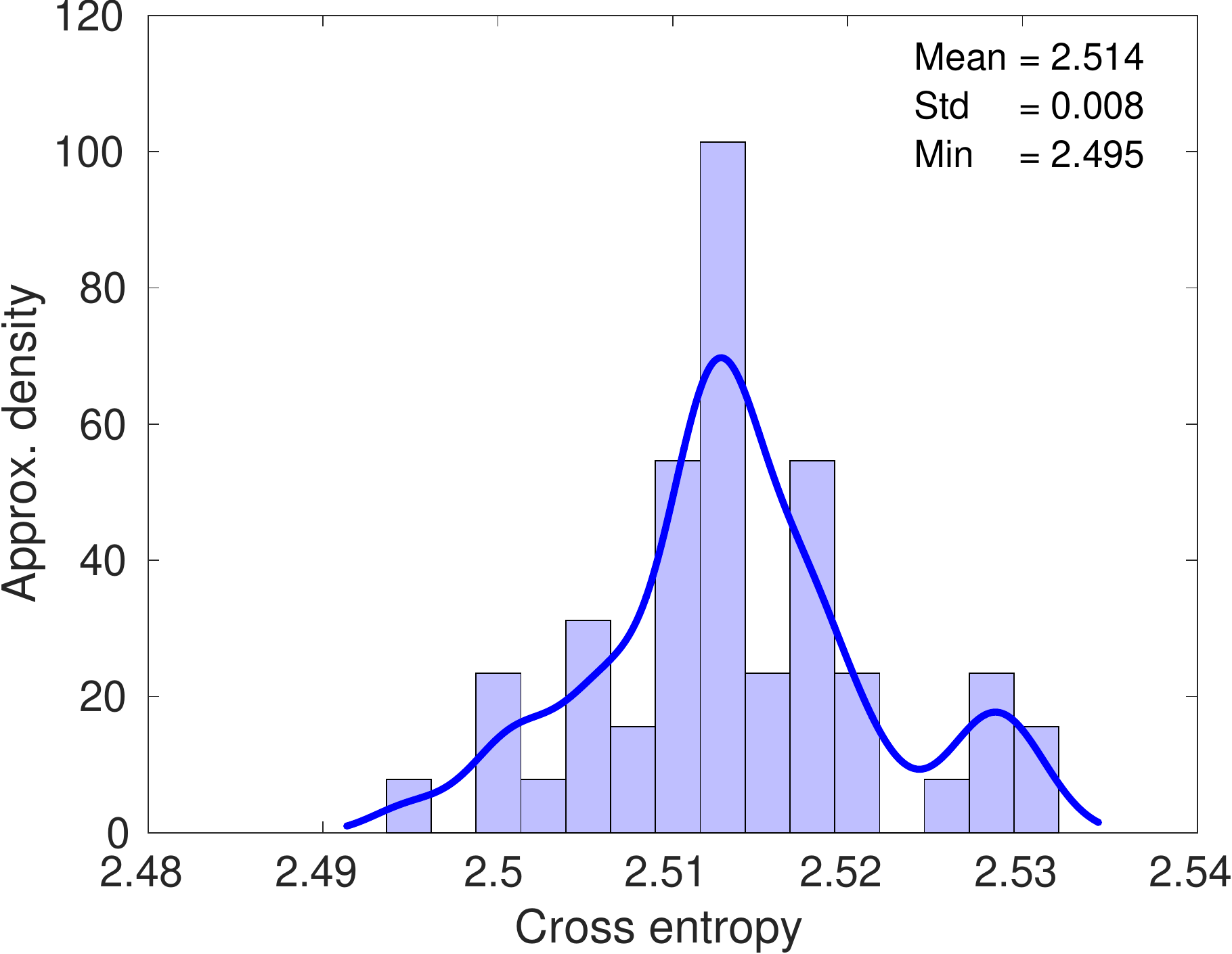}&
\includegraphics[width=0.305\textwidth]{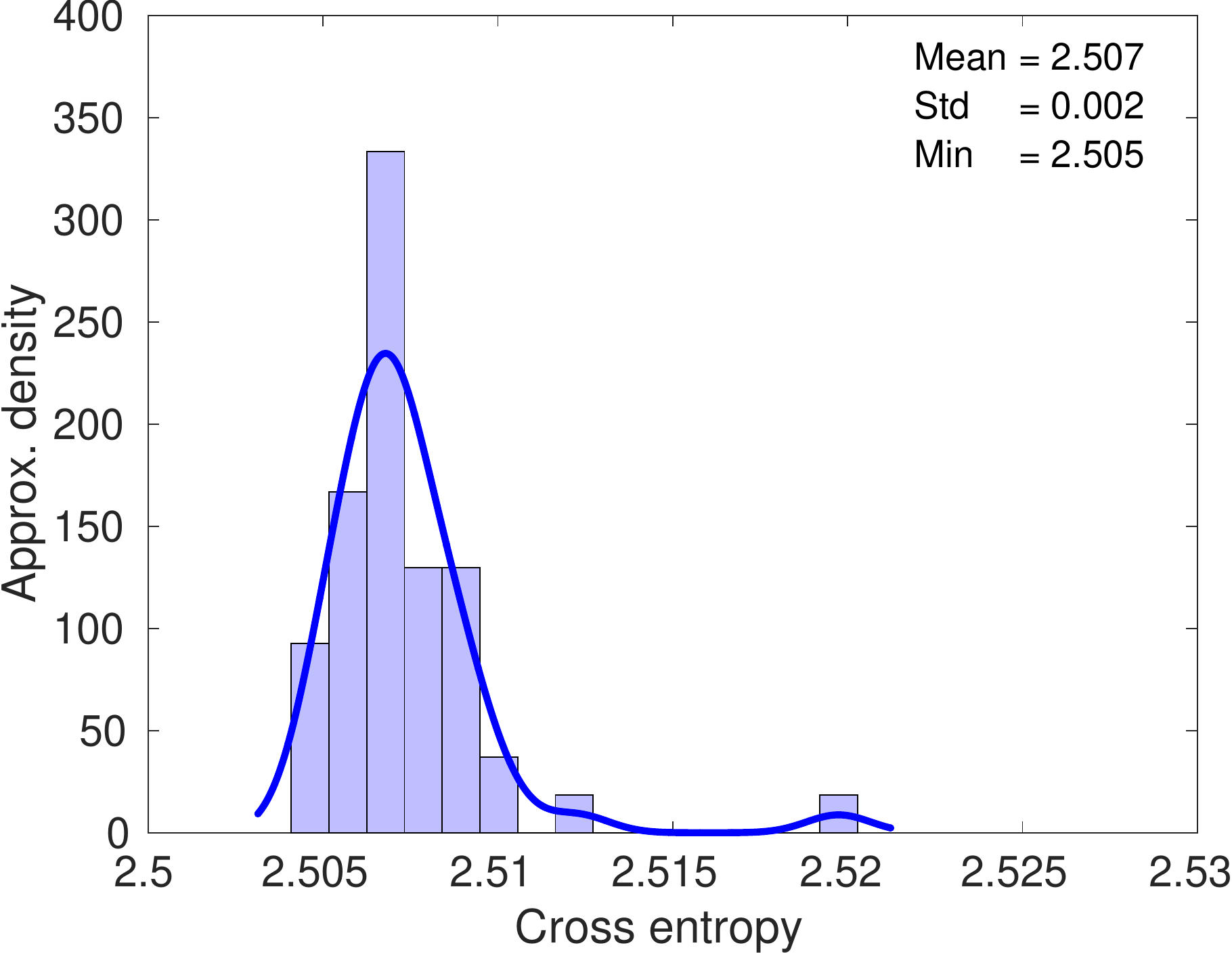}\\
({\bf{a}})&
({\bf{b}})&
({\bf{c}})\\[5pt]
\includegraphics[width=0.305\textwidth]{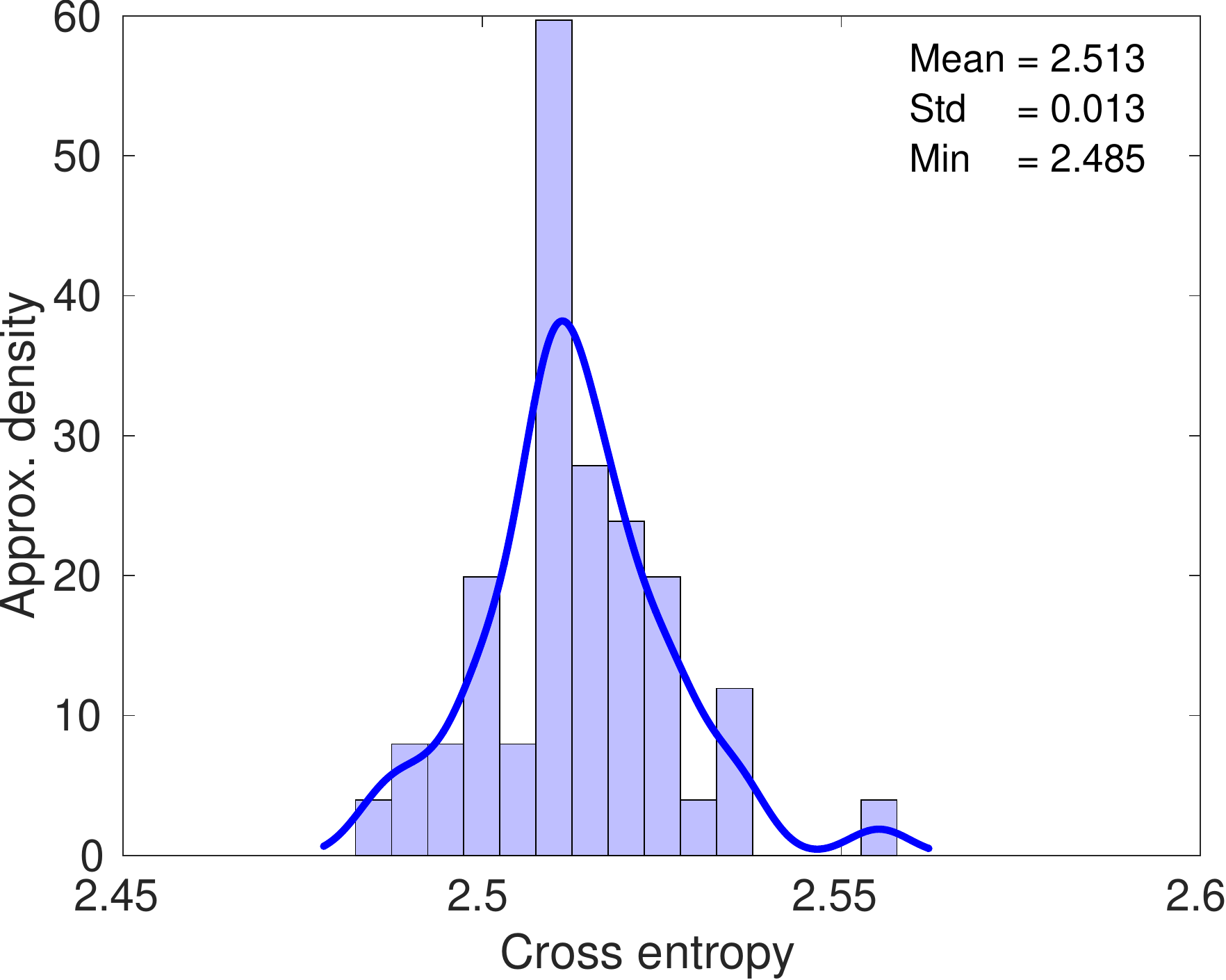}&
\includegraphics[width=0.305\textwidth]{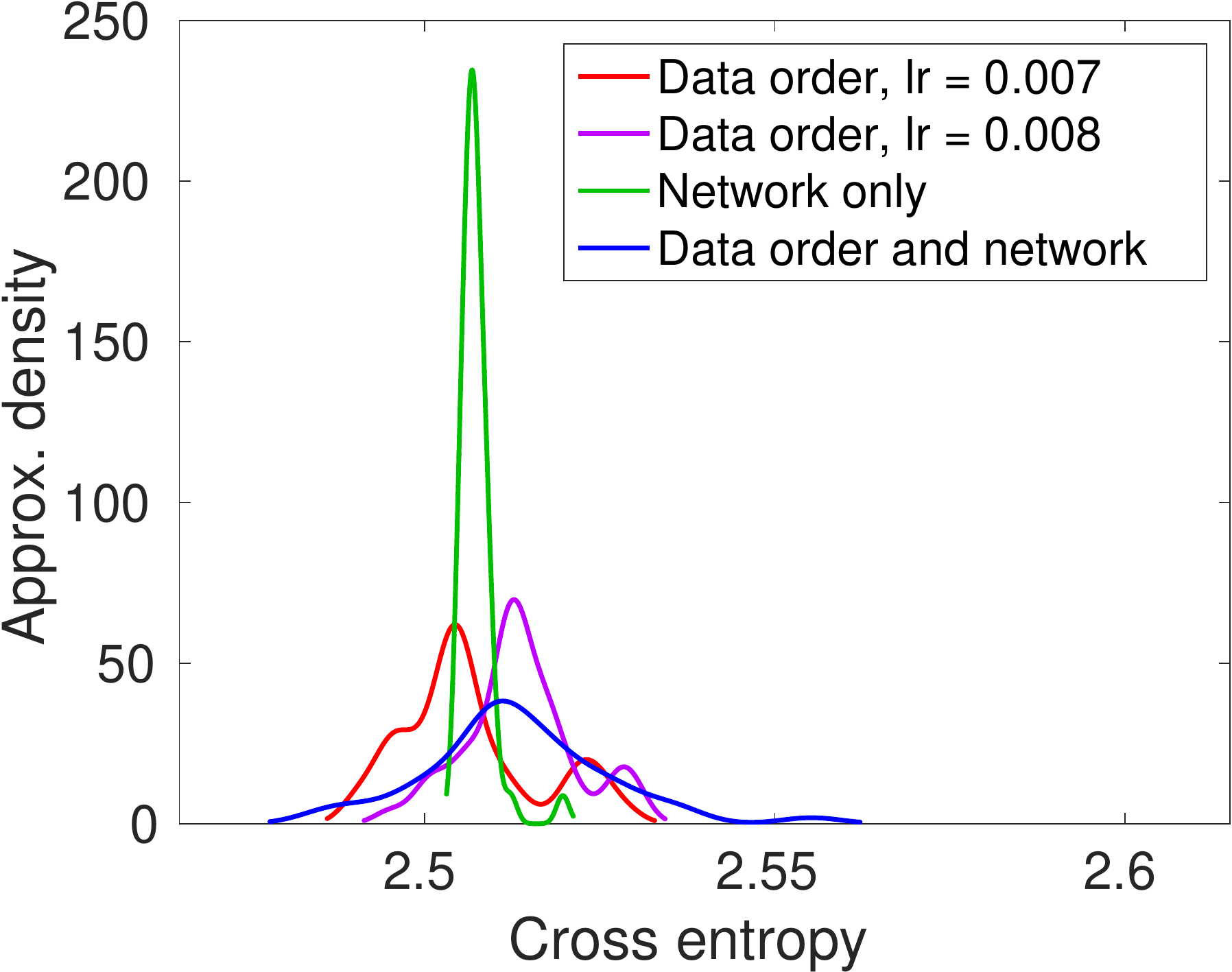}&
\includegraphics[width=0.305\textwidth]{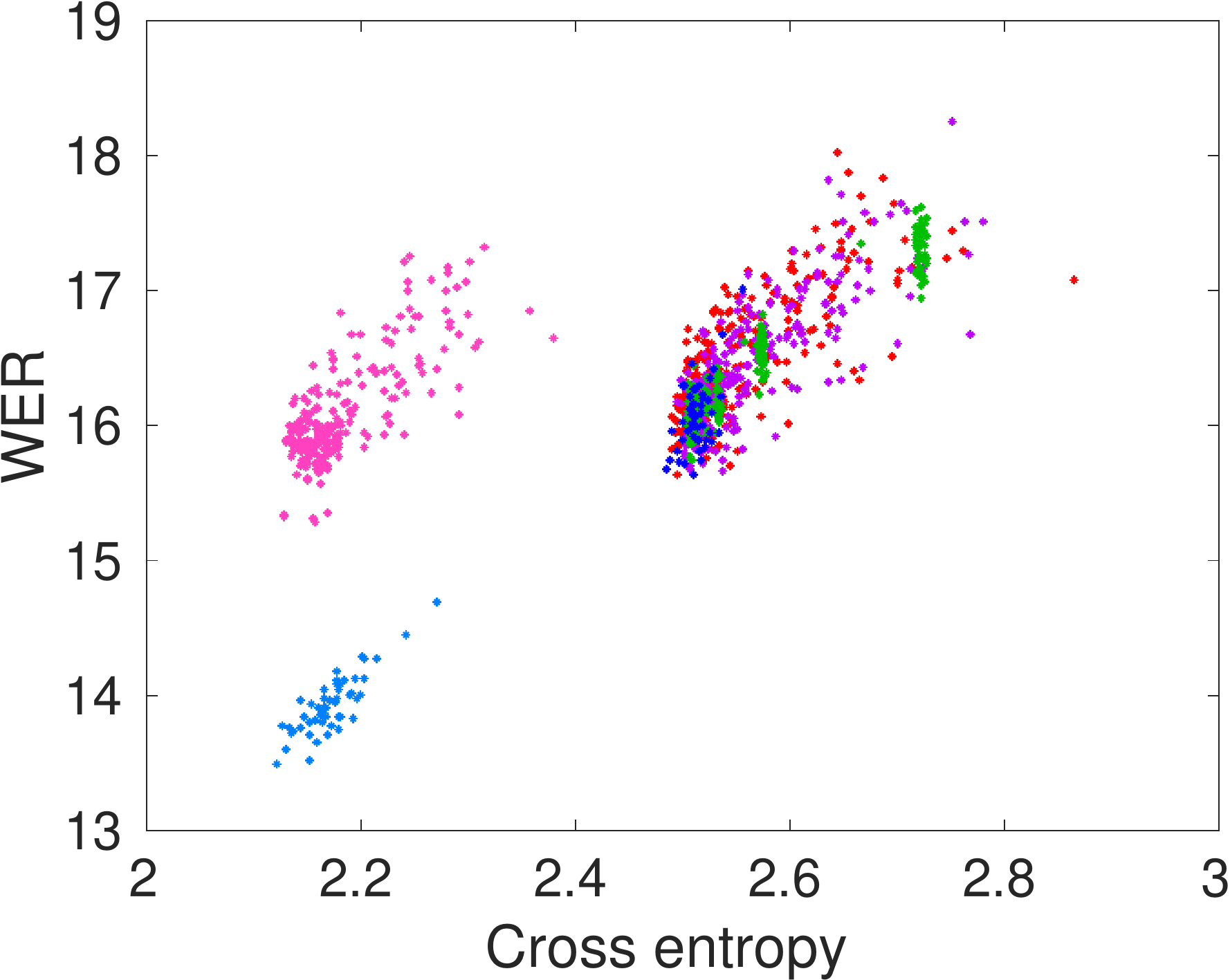}%
\begin{picture}(0,0)(0,0)
\put(-50,64){\tiny{BN-50, DNN}}
\put(-102,31){\tiny{BN-400, DNN}}
\put(-115,91){\tiny{BN-50, CNN}}
\end{picture}\\
({\bf{d}})&
({\bf{e}})&
({\bf{f}})
\end{tabular}
\caption{Distribution of the cross-entropy values after 30 epochs of
  SGD training for 50 different DNNs instances using the BN50 dataset
  with (a,b) varying mini-batch order and fixed initial network
  parameters for learning rates 0.007 and 0.008; (c) varying initial
  parameters and fixed mini-batch order, learning rate 0.008; (d)
  varying data order and parameters for learning rate 0.008; (e) a
  summary of the curves in plots (a)--(d); and (f) word-error rates
  plotted against the cross entropy for each of these systems as well
  as those for experiments on BN50 CNN and BN400 DNN.
}\label{Fig:CrossEntropy}
\end{figure*}

Most of the training is done by minimizing a cross-entropy loss
function using SGD in combination with the newbob schedule in which
the learning rate is halved whenever there is an insufficient decrease
in the held-out loss. In addition, we reset the network weights to
those from the previous epoch whenever the held-out loss
increases. For the DNN we use layerwise pre-training, each for one epoch~\cite{SAI2012BRa}. For the BN50-DNN task we also use
Hessian-free sequence training \cite{KIN2009a, MAR2010a, KIN2012SSa}.

\subsection{Results}

In the first experiment we study the empirical distribution of
cross entropy values evaluated on the hold-out set by training 50 DNN
networks for the BN50 task with different random seeds for the initial
network parameters, while fixing the order of the mini batches. The
results obtained with initial learning rates set to 0.007 and 0.008
are plotted in Figures~\ref{Fig:CrossEntropy}(a,b) as both histograms
and kernel density estimates using a Gaussian kernel (solid
line). The difference in learning rate is not very large and the
distributions are fairly similar except that the mean for the
experiments with learning rate 0.007 is somewhat smaller than that of
0.008. In the next experiment we use a fixed random initialized of the
network parameters, and then vary the random seed used for the batch
randomization. With a learning rate of 0.008 this gives the
distribution shown in Figure~\ref{Fig:CrossEntropy}(c). As seen from
the plot and as evident from the standard deviation, the distribution
is tightly concentrated around its mean with the exception of one
outlier. Next, we randomize both the initial parameters as well as the
data order, while keeping the initial learning rate at 0.008. The
results in Figure~\ref{Fig:CrossEntropy}(d) show that the standard
deviation grows and exceeds the sum of those in
Figures~\ref{Fig:CrossEntropy}(b) and (c). For reference we plot all
kernel density estimates in Figure~\ref{Fig:CrossEntropy}. From this
plot it is clear just how localized the distribution of cross entropy
values is when keeping the data order fixed and changing only the
initial network parameters. This suggests that the data order is more
important than the initial parameters in determining the final
result. More experiments are needed to see if this applies only for
this task, or whether it holds more generally.

\begin{figure}[h*]
\centering
\includegraphics[width=0.40\textwidth]{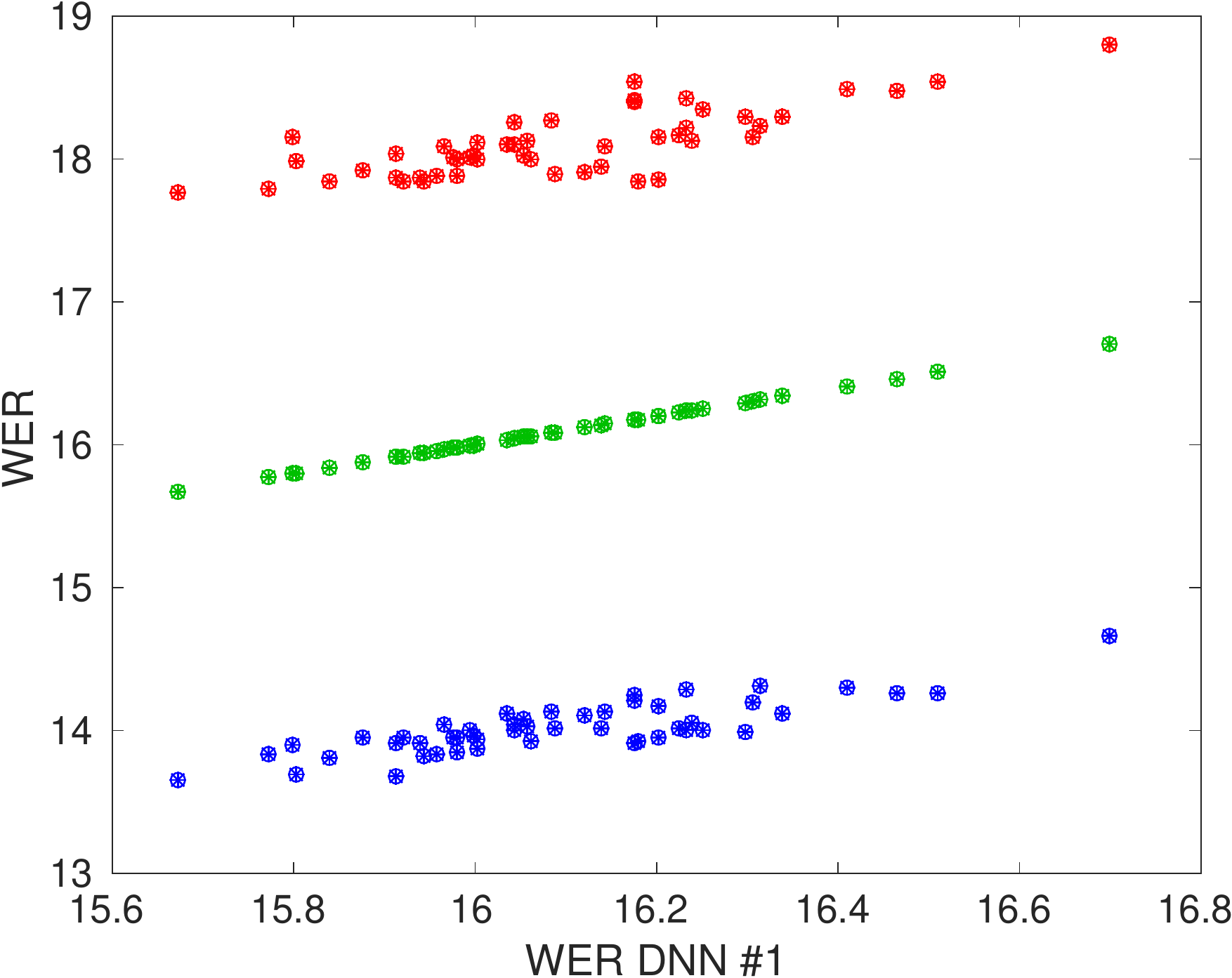}
\caption{The word-error rates depend on the language model and
  evaluation set: (top) standard LM, evaluated on RT-04, (middle)
  standard LM, Dev-04f, and (bottom) large LM, Dev-04f. Results are
  plotted against the WER results in the middle.}\label{Fig:WERxWER}
\end{figure}

As mentioned in the introduction, the goal of ASR is to obtain a small
word-error rate. In addition to the acoustic model, the word-error
rate also depends on the decoder settings, the language model, and the
relative weights of the language and acoustic models for computing
likelihoods. In Figure~\ref{Fig:CrossEntropy}(f) we plot the
word-error rates for the above four settings against the
cross-entropy. While the two are somewhat correlated, it does not hold
that models with a lower cross entropy necessarily have a word-error
rate as well. We also trained 50 instances of the BN400 DNN task with
varying initial parameters and data orders. The results of these
experiments appear in the bottom left corner of the figure. Finally,
we trained 50 CNN instances on BN50, again with different initial
parameters and data order. Even though the cross entropy is closer to
the values obtained with the BN400 DNNs, the word-error rate only
seems to improve marginally over the BN50 DNNs. It is not clear what
causes this disparity, but it does clearly show the discrepancy
between the training objective and the evaluation metric. To see how
the performance of the models changes with different evaluation sets
and language models, we plot in Figure~\ref{Fig:WERxWER} the results
of three different decodes against the results from the BN50 DNN
trained with learning rate 0.008, standard language model and
evaluated on Dev04f. Overall the relative model accuracy roughly
remain the same across evaluation sets and language models. In all
three settings the best models remains the best and likewise for the
worst model.


\begin{figure*}[t]
\centering
\begin{tabular}{ccc}
\includegraphics[width=0.295\textwidth]{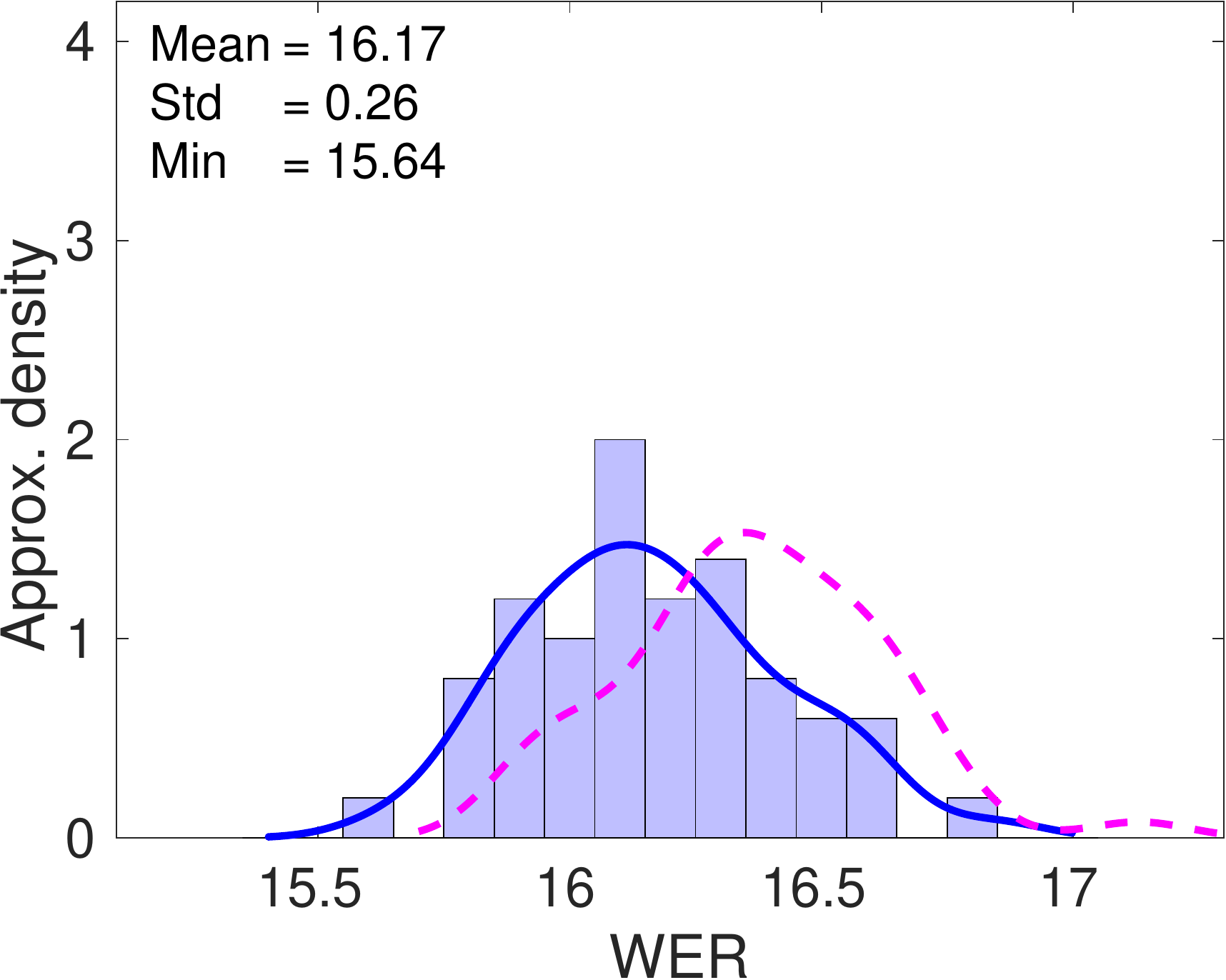} &
\includegraphics[width=0.295\textwidth]{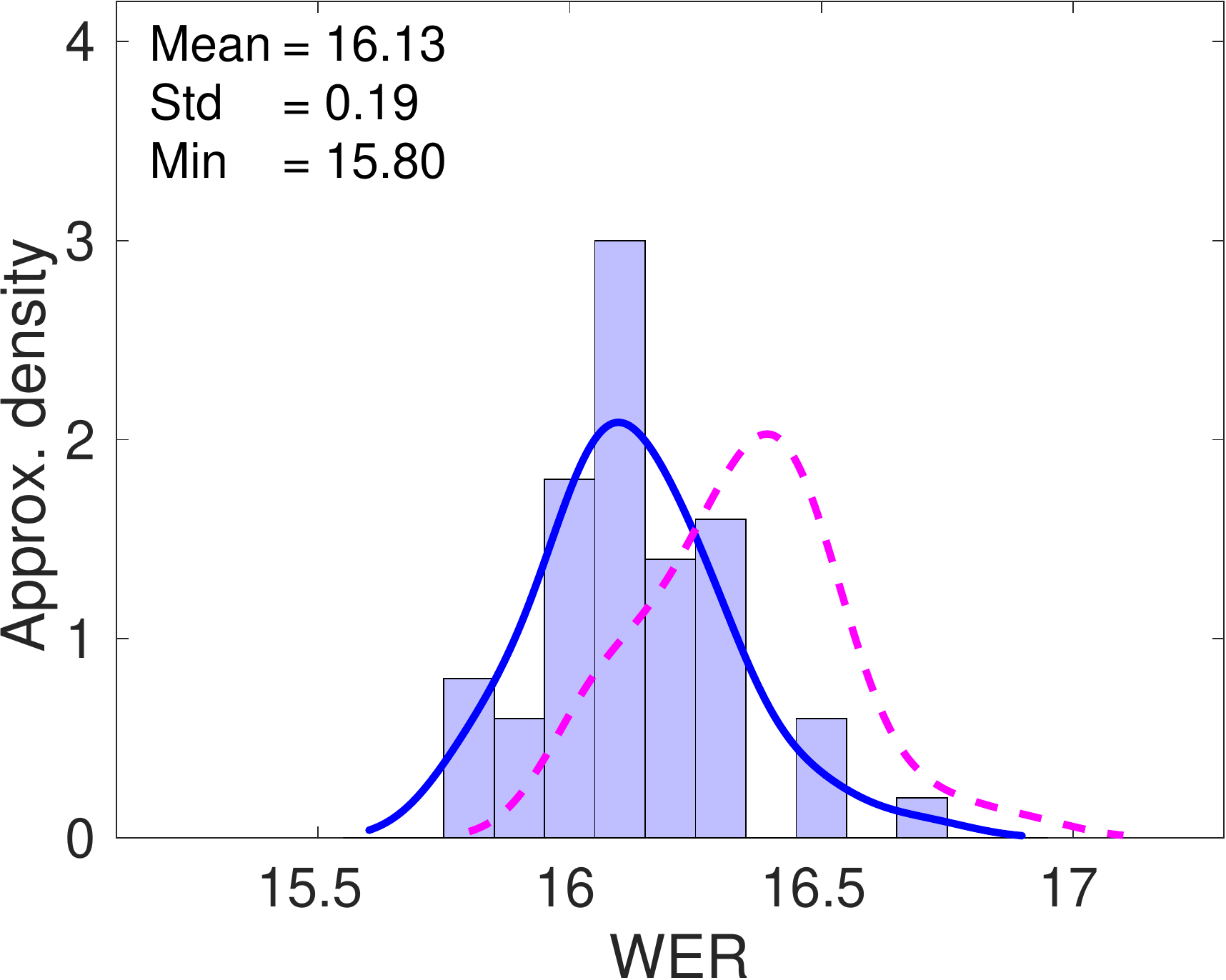} &
\includegraphics[width=0.295\textwidth]{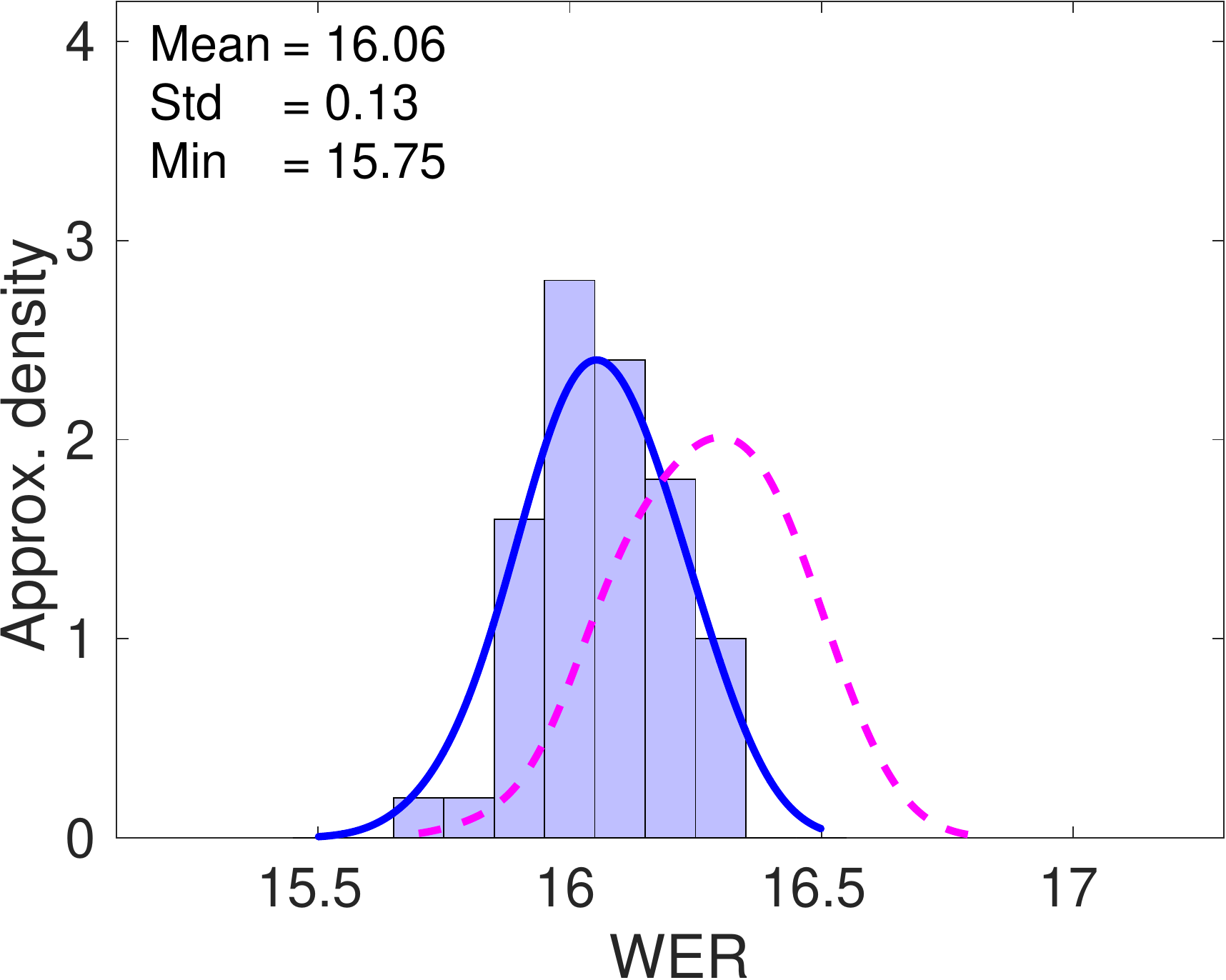} \\
{\small ({\bf{a}})  {\sc{bn{\scriptsize 50}}-dnn}, \footnotesize data
  order, lr = 0.007} &
{\small ({\bf{b}})  {\sc{bn{\scriptsize 50}}-dnn}, \footnotesize data
  order, lr = 0.008} &
{\small ({\bf{c}})  {\sc{bn{\scriptsize 50}}-dnn}, \footnotesize network only}\\[5pt]
\includegraphics[width=0.295\textwidth]{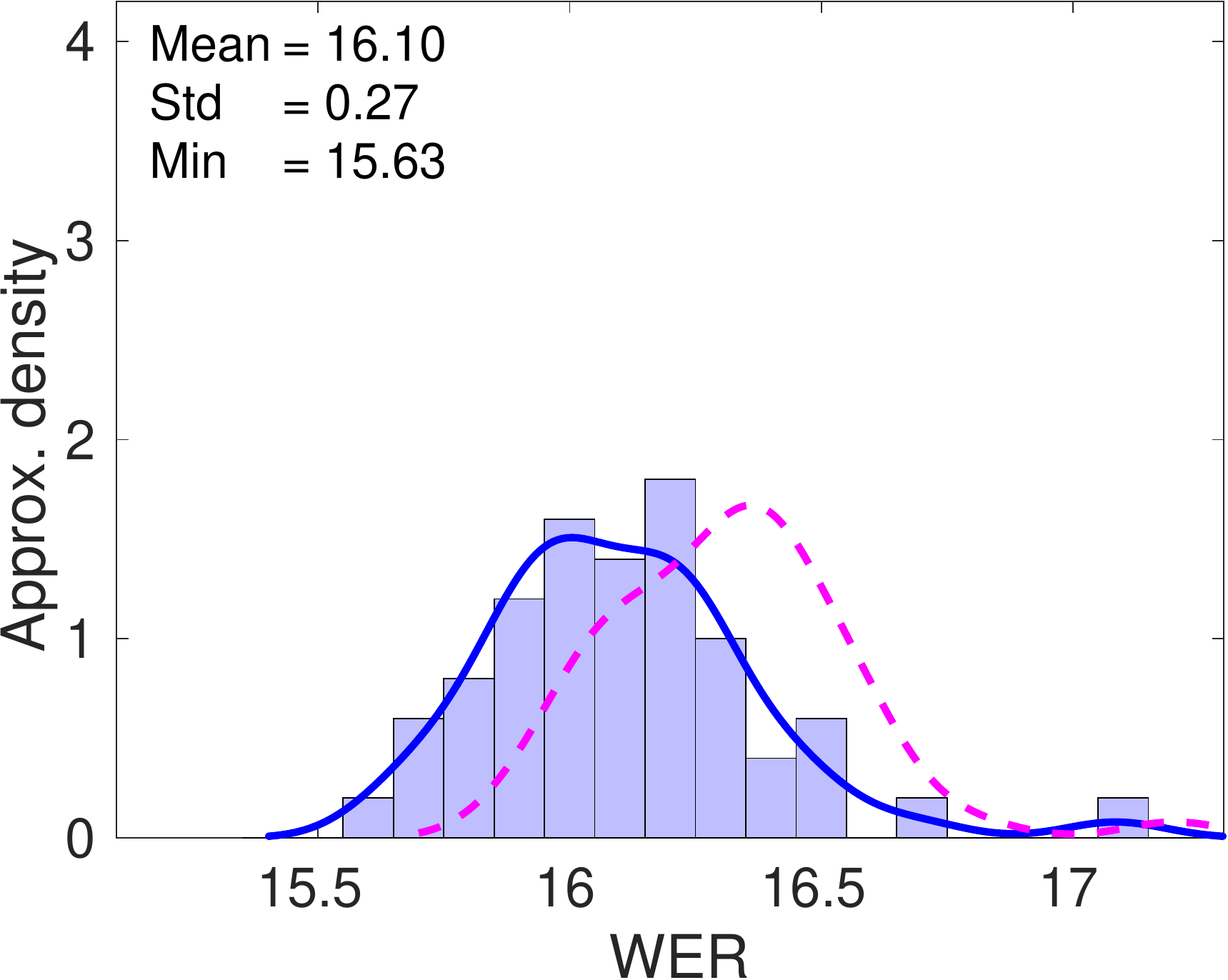} &
\includegraphics[width=0.295\textwidth]{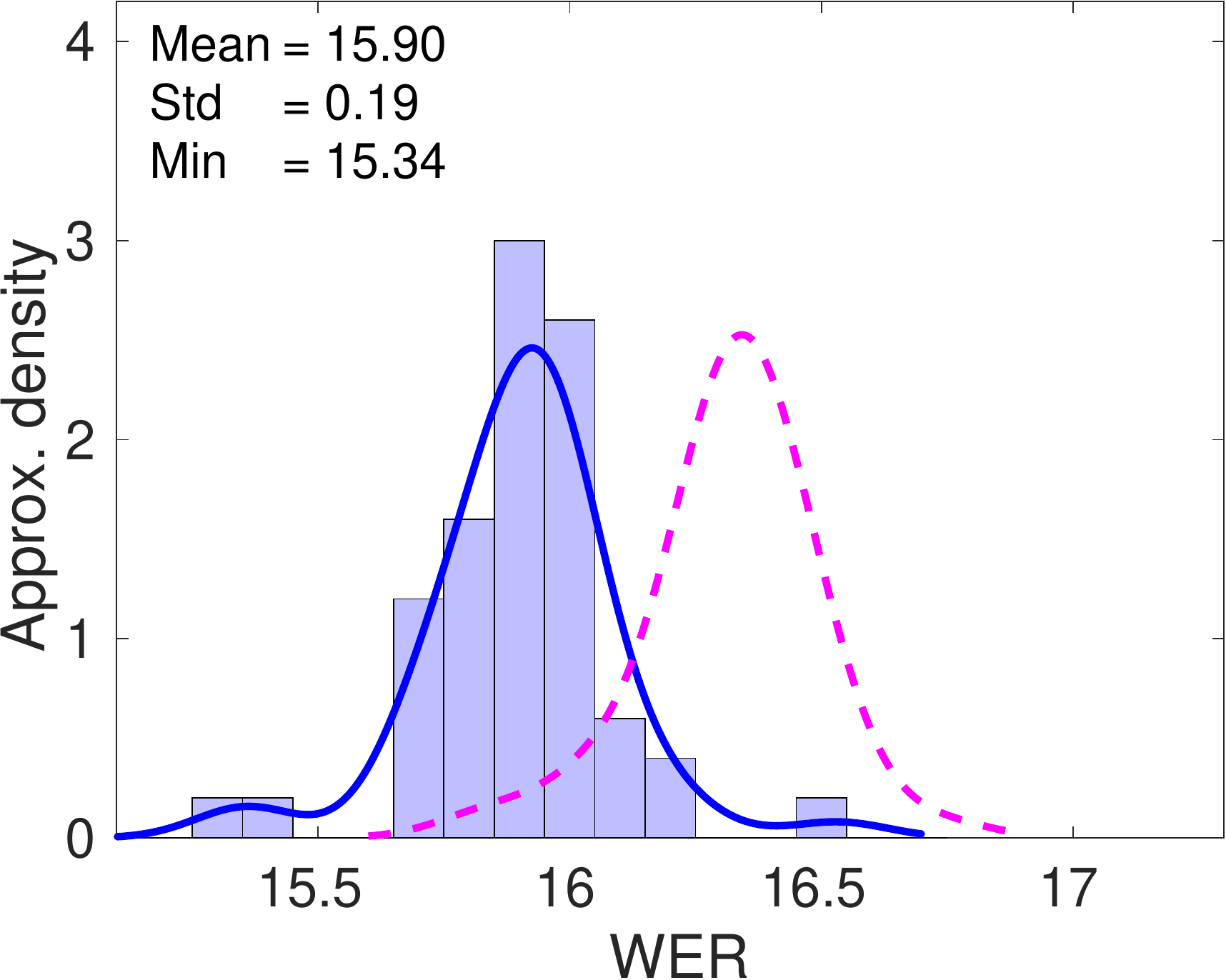} &
\includegraphics[width=0.295\textwidth]{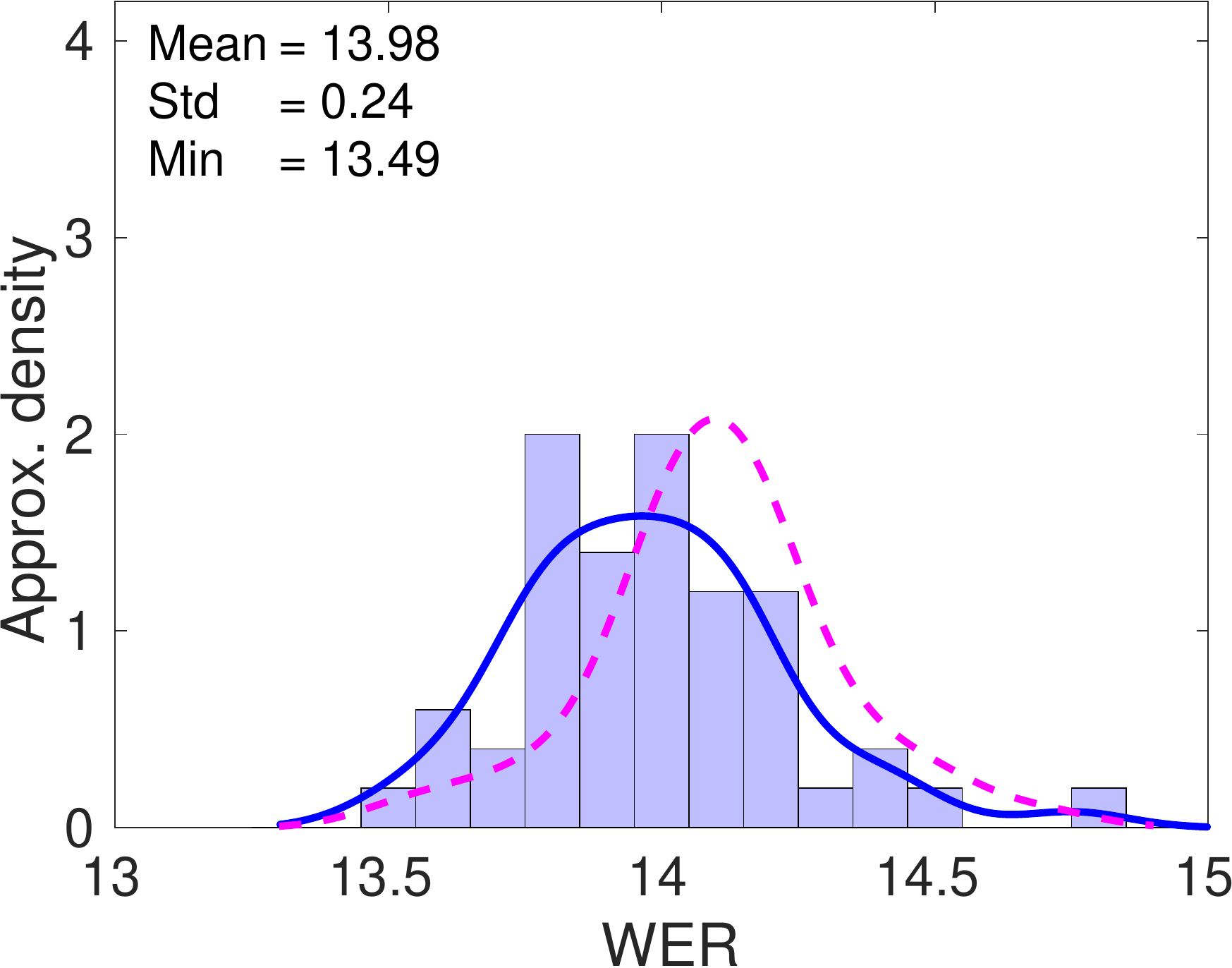} \\
{\small ({\bf{d}})  {\sc{bn{\scriptsize 50}}-dnn}, \footnotesize data order \& network} &
{\small ({\bf{f}})  {\sc{bn{\scriptsize 50}}-cnn}, \footnotesize data order \& network} &
{\small ({\bf{e}})  {\sc{bn{\scriptsize 400}}-dnn}, \footnotesize data order \& network} \\[5pt]
\includegraphics[width=0.295\textwidth]{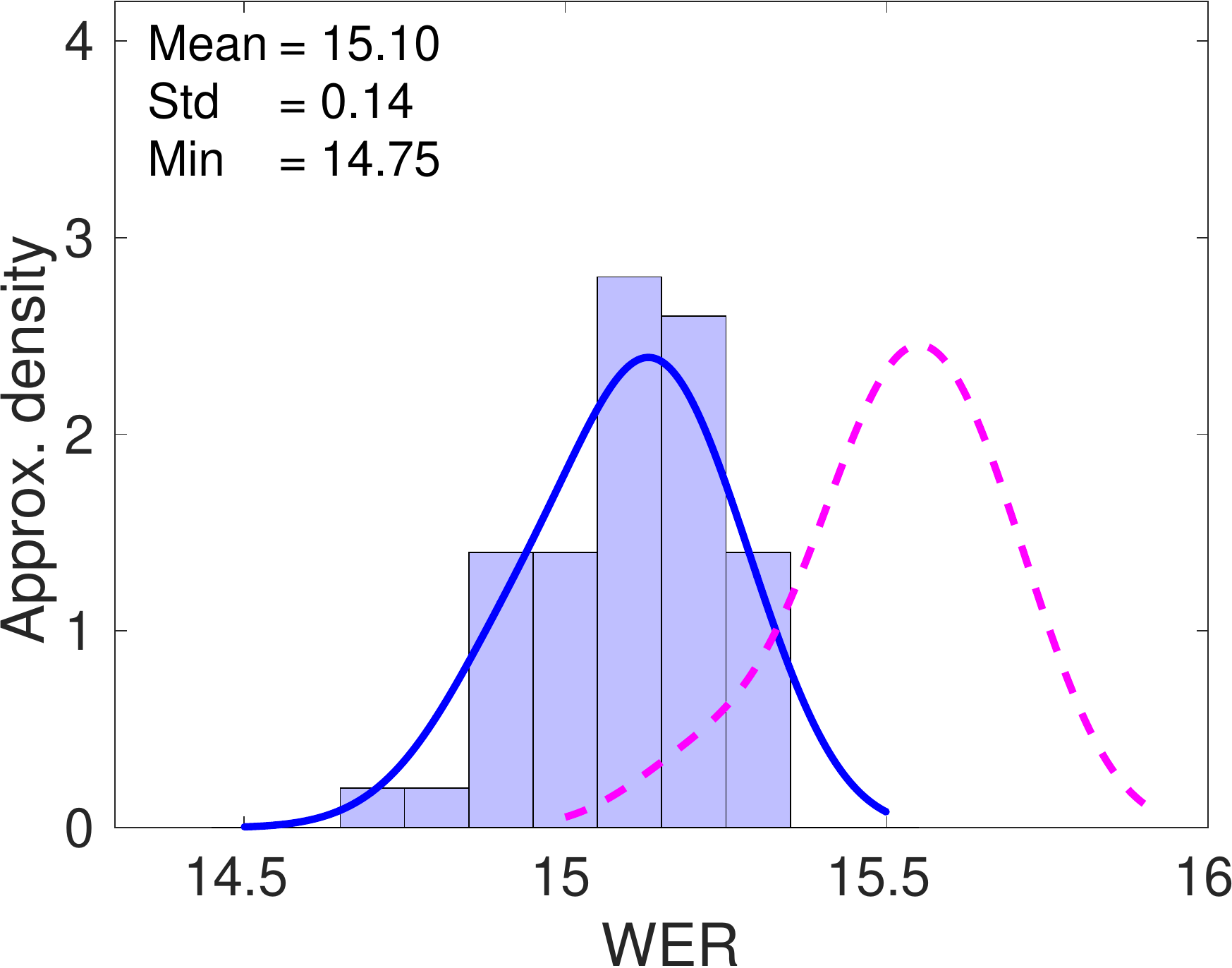} &
\includegraphics[width=0.295\textwidth]{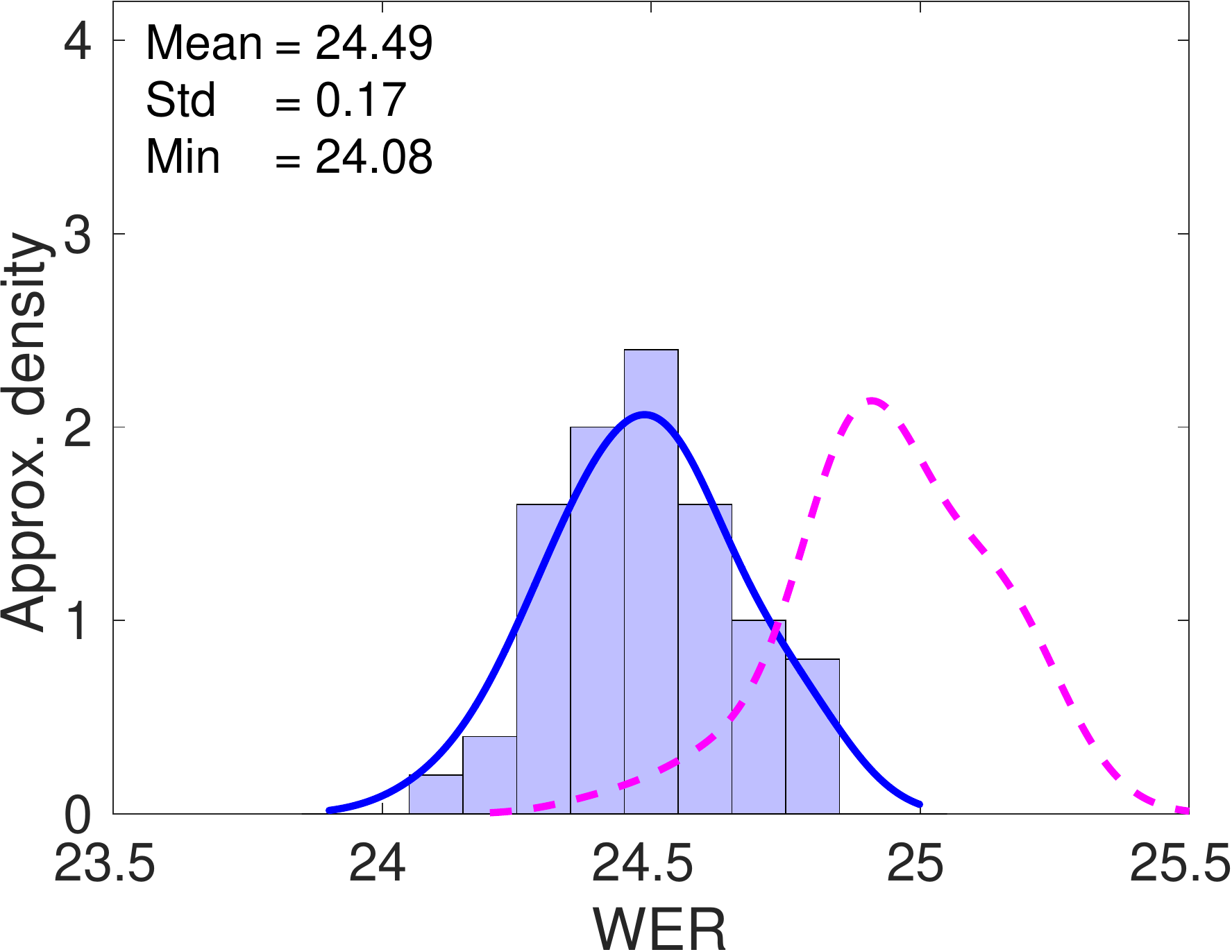} &
\includegraphics[width=0.295\textwidth]{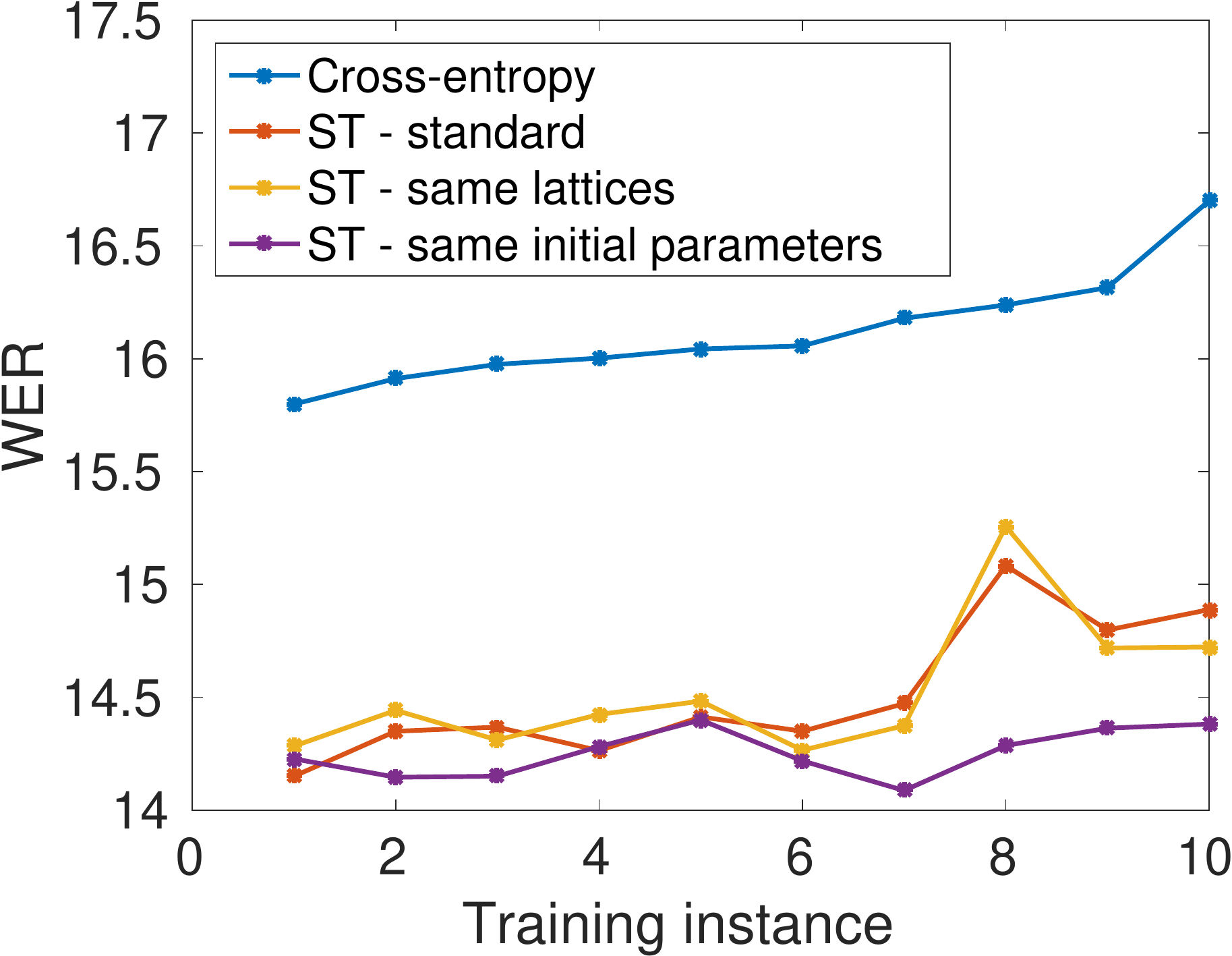} \\
{\small ({\bf{g}})  {\sc{swb-dnn}}, \footnotesize Hub5} &
{\small ({\bf{h}})  {\sc{swb-dnn}}, \footnotesize CallHome} &
{\small ({\bf{i}})  {\sc{bn{\scriptsize 50}}-dnn}, \footnotesize
  sequence training}\\[5pt]
\end{tabular}
\caption{Result obtained for different datasets and neural network
  types, as indicated next to the labels. Plots (a)--(d) show the
  results for BN50 DNN setups; (e) gives the results for the BN50 CNN
  system with varying data order and initial network parameters; (f)
  gives similar result for a BN400 DNN setup; (g,h) plot the results
  for the 300h switchboard DNN system when evaluated respectively on
  the Hub5 and CallHome datasets. The underlying acoustic for these
  two are the same and each of the 50 instances is trained with
  different random seeds for the data order and the initial network
  parameters. Finally, (i) gives the results obtained with
  sequence training.}\label{Fig:WERHist}
\end{figure*}

In Figures~\ref{Fig:WERHist}(a)--(d) we plot the empirical
distribution of the WER for the four BN50 DNN experiments given in
Figure~\ref{Fig:CrossEntropy}(e). The blue histogram and kernel
density estimation summarize the results obtained when using the
optimal weight on the acoustic model relative to the language
model. The dashed purple line shows the distribution that results when
changing the acoustic weight to a sub-optimal value. The mean and
standard deviation in WER values for the different setups reflect
the relative values of the cross entropy values in
Figures~\ref{Fig:CrossEntropy}(a)--(d). The largest WER range of 15.6 to
17.1 is obtained when both the initial parameters and data order vary.
Among the BN50 configurations tried, the results for the BN50 CNN
setup in Figure~\ref{Fig:WERHist}(e) have the smallest mean and
minimum WER. In addition, the standard deviation is small compared to
the equivalent DNN in Figure~\ref{Fig:WERHist}(d).

Increasing the amount of training data gives a large reduction in WER
with the average going down from 16.1 to 13.98, as shown in
Figure~\ref{Fig:WERHist}(e). The difference between the best and worst
WER remains around 1.3 percent absolute, but the ratio between the
standard deviation and the mean remains nearly the same being only
slightly higher for the BN400 setup. Figures~\ref{Fig:WERHist}(g) and
(h) give the distribution of WER for the 300-hour Switchboard system
evaluated on the Hub5 and CallHome datasets. The standard deviation on
these experiments is smaller than those for the different Broadcast
News experiments.

So far we have only considered experiments based on cross-entropy
minimization using SGD. The best results in speech, see for example
\cite{SAO2015KRPa}, are obtained using sequence-level discriminative
training techniques. For our experiments we minimize the state-based
minimum Bayes risk (MBR) objective \cite{KIN2009a,KIN2012SSa} using
Hessian-free optimization \cite{MAR2010a}. As a starting point we took
ten networks from the BN50 DNN setup in which both the data order and
initial network parameters change and denote then by $n_1$ through
$n_{10}$ in decreasing order of performance. For each of these
networks we generated the corresponding enumerator and denominator
lattices using a unigram language model, thus obtaining $\ell_1$
through $\ell_{10}$. We then applied sequence training to three
settings: (1) Initial network $n_k$ and lattices $\ell_k$; (2) initial
network $n_k$ and fixed lattice $\ell_1$; and (3) fixed initial
network $n_1$ and lattices $\ell_k$. The resulting WERs are plotted in
Figure~\ref{Fig:WERHist}(i) and connected by lines for clarity. All
sequence-training results can be seen to improve substantially over
the cross-entropy results. Interestingly, the starting point seems to
be much more important than the network quality used to generate the
lattices; initializing the network parameters with those of the best
BN50 DNN cross-entropy system gives the best results among the three
setups.

\section{Discussion}\label{Sec:Discussion}

When evaluating the performance of a new method there are two aspects
that need to be taken into consideration. The first one concerns the
statistical significance of results obtained for individual
models. This includes techniques such as McNemar's and matched-pair
tests, as advocated in \cite{GIL1989Ca}. The second aspects goes
beyond individual models and concerns the performance of methods as a
whole. Currently, methods are evaluated by comparing with a baseline
the very best model obtained after careful and often extensive fine
tuning. The baseline may itself be a highly refined result reported in
the literature, but equally often consist of a single run of a
standard DNN or CNN system. Aside from the question of whether the two
models are significantly different is the more important question of
how meaningful it is to compare competing methods based on only one
pair of models per dataset. As a though experiment, consider the
situation where we compare the performance of the BN50 DNN from
Figure~\ref{Fig:WERHist}(a) against itself. In particular, we consider
the distribution in Figure~\ref{Fig:WERHist}(a) and compute the
probability that, based on a matched-pair test with given confidence
level, at least one of $n$ randomly sampled models is significantly
better than a randomly sampled baseline. This gives:
\begin{center}
\begin{tabular}{lrrrrr}
Conf. & $n=1$ & $n=2$ & $n=5$ & $n=10$ & $n=20$\\
95.0\% & 35.0 & 54.7 & 81.9 & 95.1 & 99.5\\
99.0\% & 22.2 & 36.6 & 61.3 & 79.3 & 92.0\\
99.9\% & 11.3 & 19.7 & 36.4 & 51.7 & 66.8\\
\end{tabular}

\end{center}
It is seen that with sufficient sampling there is a high probability
of finding a model that improves significantly over the baseline, and
we would therefore conclude that the method improves upon itself! In
practice a similar setting may arise when a high degree of fine tuning
amounts not so much to finding the optimal parameters but rather to
sampling repeatedly from similar distributions, thereby increasing the
chance of eventually finding a good model instance. We see that even
if a model obtained with a proposed method is significantly better
than the baseline, it does not automatically allow us to conclude that
the method as a whole is better than that used for the baseline.

When the training distribution of two methods were known (either based
on an optimal set of hyperparameters, or on some distribution over
these parameters) we could look at principled ways of comparing the
performance. We could for example

\begin{enumerate}
\item Compare the best possible results; this provides a best-case
  scenario and provided that the baseline comes from a highly tuned
  system, this is what is commonly used. Note that this approach is
  biased in favor of the proposed method if the baseline is not
  extensively tuned (as illustrated above).
\item Compare the mean results; this says something about the
  performance of an average training run;
\item Determine the probability that a randomly sampled models of
  one method is better than that the other.
\end{enumerate}

Of course, the entire distribution is not known and even obtaining a
good approximation may require excessive sampling. For a comparison
methodology to be practicable we require that it can be evaluated
based on only a small number of samples, which is even more
challenging if the underlying distribution is non-parametric. The
performance of the first approach really depends on the probability
mass contained in the lower tail; if this is small it will be very
difficult to sample. When comparing the mean care needs to be taken;
suppose that the means are identical. Then in terms of method
performance it may be desirable to have a stable method with little
variance. On the other hand, when it comes to obtaining a model for
practical use, a larger variance will make it easier to sample the
best model (here the goal really is to obtain the best possible model
instance). For the third example criterion, we could sample pairs of
models and apply a pairwise test and assign 1 for a win, $1/2$ for a
draw, and 0 for losing. The sum of these values gives the relative
performance of the methods, which can be quantified by testing against
a binomial distribution with suitable $p$ values. Alternatively, a
pairwise test over the results for each method combined can be done.

\section{Conclusions}\label{Sec:Conclusions}

In the literature it is still common practice to report results
without any information on the variance; see for example
\cite{SAI2013KMDa,SAI2015VSSa,ZHA2014LGa,LI2015MZGa}. There are of course some
exceptions such as \cite{GRA2013JMa,MOO2015CLSa}, but even these
papers only give variance on phone-error rate and not on WER. The
improvements in the aforementioned papers may very well be meaningful
and represent a significant improvement on the baseline method, but
the problem is that based on individual numbers we simple cannot be
sure. There is admittedly some additional difficulty in systems that
rely on a large number of training stages. The sequence training setup
discussed in the previous section is a simple example, but much more
involved processing approaches in which each step is highly tuned
exist \cite{SAO2015KRPa}. Nevertheless, more insight into the result
distributions for different settings and algorithms, and their
incorporation into performance evaluation is much needed.

  \eightpt
  \bibliographystyle{IEEEtran}
  \bibliography{bibliography}

\begin{thebibliography}{10}
\providecommand{\url}[1]{#1}
\csname url@samestyle\endcsname
\providecommand{\newblock}{\relax}
\providecommand{\bibinfo}[2]{#2}
\providecommand{\BIBentrySTDinterwordspacing}{\spaceskip=0pt\relax}
\providecommand{\BIBentryALTinterwordstretchfactor}{4}
\providecommand{\BIBentryALTinterwordspacing}{\spaceskip=\fontdimen2\font plus
\BIBentryALTinterwordstretchfactor\fontdimen3\font minus
  \fontdimen4\font\relax}
\providecommand{\BIBforeignlanguage}[2]{{%
\expandafter\ifx\csname l@#1\endcsname\relax
\typeout{** WARNING: IEEEtran.bst: No hyphenation pattern has been}%
\typeout{** loaded for the language `#1'. Using the pattern for}%
\typeout{** the default language instead.}%
\else
\language=\csname l@#1\endcsname
\fi
#2}}
\providecommand{\BIBdecl}{\relax}
\BIBdecl

\bibitem{SEI2011LYa}
F.~Seide, G.~Li, and D.~Yu, ``Conversational speech transcription using
  context-dependent deep neural networks,'' in \emph{Interspeech 2011}.\hskip
  1em plus 0.5em minus 0.4em\relax International Speech Communication
  Association, August 2011.

\bibitem{CHO2014HMAa}
A.~Choromanska, M.~Henaff, M.~Mathieu, G.~B. Arous, and Y.~Le{C}un, ``The loss
  surfaces of multilayer networks,'' arXiv:1412.0233, 2014.

\bibitem{PAS2014DGBa}
R.~Pascanu, Y.~N. Dauphin, S.~Ganguli, and Y.~Bengio, ``On the saddle point
  problem for non-convex optimization,'' arXiv:1405.4604, 2014.

\bibitem{PIN2009DDCa}
N.~Pinto, D.~Doukhan, J.~J. Di{C}arlo, and D.~D. Cox, ``A high-throughput
  screening approach to discovering good forms of biologically inspired visual
  representation,'' \emph{PLoS Comput Biol}, vol.~5, no.~11, p. e100579, 2009.

\bibitem{HIN2015VDa}
G.~Hinton, O.~Vinyals, and J.~Dean, ``Distilling the knowledge in a neural
  network,'' arXiv:1503.02531, 2015.

\bibitem{FIS1997GPFa}
J.~Fiscus, J.~Garofolo, M.~Przybocki, W.~Fisher, and D.~Pallett, ``1997
  {E}nglish {B}roadcast {N}ews {S}peech ({HUB4}) {LDC98S71},'' 1998, linguistic
  Data Consortium.

\bibitem{KIN2009a}
B.~Kingsbury, ``Lattice-based optimization of sequence classification criteria
  for neural-network acoustic modeling,'' in \emph{Proceedings of ICASSP},
  2009, pp. 3761--3764.

\bibitem{GOD1997Ha}
J.~Godfrey and E.~Holliman, ``Switchboard-1 {R}elease 2 {LDC97S62},'' 1997,
  {L}inguistic Data Consortium, Philadelphia.

\bibitem{LIU1995MNPa}
F.-H. Liu, M.~Monkowski, M.~Novak, M.~Padmanabhan, M.~Picheny, and P.~S. Rao,
  ``Performance of the {IBM} {LVCSR} system on the {S}witchboard corpus,'' in
  \emph{Speech Research Symposium}, 1995, p. 189.

\bibitem{SAO2015KRPa}
G.~Saon, H.-K.~J. Kuo, S.~Rennie, and M.~Picheny, ``The {IBM} 2015 {E}nglish
  conversational telephone speech recognition system,'' in \emph{Proceedings of
  Interspeech}, 2015.

\bibitem{SAI2012BRa}
T.~N. Sainath, B.~Kingsbury, and B.~Ramabhadran, ``Improving training time of
  deep belief networks through hybrid pre-training and larger batch sizes,'' in
  \emph{Proceedings of the NIPS Workshop on Log-linear Models}, 2012.

\bibitem{MAR2010a}
J.~Martens, ``Deep learning via {H}essian-free optimization,'' in
  \emph{Proceedings of the 27th International Conference on Machine Learning
  (ICML)}, 2010.

\bibitem{KIN2012SSa}
B.~Kingsbury, T.~Sainath, and H.~Soltau, ``Scalable minimum {B}ayes risk
  training of deep neural network acoustic models using distributed
  {H}essian-free optimization,'' in \emph{Proceedings of Interspeech}, 2012.

\bibitem{GIL1989Ca}
L.~Gilick and S.~J. Cox, ``Some statistical issues in the comparison of speech
  recognition algorithms,'' in \emph{Proceedings of the 1989 International
  Conference on Acoustics, Speech, and Signal Processing (ICASSP)}, vol.~1, May
  1989, pp. 532--535.

\bibitem{SAI2013KMDa}
T.~N. Sainath, B.~Kingsbury, A.-R. Mohamed, G.~E. Dahl, G.~Saon, H.~Soltau,
  T.~Beran, A.~Y. Aravkin, and B.~Ramabhadran, ``Improvements to deep
  convolutional neural networks for {LVCSR},'' in \emph{2013 IEEE Workshop on
  Automatic Speech Recognition and Understand (ASRU)}, December 2013, pp.
  315--320.

\bibitem{SAI2015VSSa}
T.~N. Sainath, O.~Vinyals, A.~Senior, and H.~Sak, ``Convolution, long
  short-term memory, fully connected deep neural networks,'' in
  \emph{Proceedings of ICASSP}, 2015, pp. 4580--4584.

\bibitem{ZHA2014LGa}
R.~Zhao, J.~Li, and Y.~Gong, ``Variable-component deep neural network for
  robust speech recognition,'' in \emph{Proceedings of Interspeech}, 2014.

\bibitem{LI2015MZGa}
\BIBentryALTinterwordspacing
J.~Li, A.~Mohamed, G.~Zweig, and Y.~Gong, ``{LSTM} {T}ime and frequency
  recurrence for automatic speech recognition,'' in \emph{IEEE Automatic Speech
  Recognition and Understanding Workshop}, December 2015. [Online]. Available:
  \url{http://research.microsoft.com/apps/pubs/default.aspx?id=259285}
\BIBentrySTDinterwordspacing

\bibitem{GRA2013JMa}
A.~Graves, N.~Jaitly, and A.-R. Mohamed, ``Hybrid speech recognition with deep
  bidirectional {LSTM},'' in \emph{2013 IEEE Workshop on Automatic Speech
  Recognition and Understanding (ASRU)}, December 2013, pp. 273--278.

\bibitem{MOO2015CLSa}
T.~Moon, H.~Choi, H.~Lee, and I.~Song, ``{RNNDrop}: A novel dropout for {RNN}s
  in {ASR},'' in \emph{2015 IEEE Workshop on Automatic Speech Recognition and
  Understand (ASRU)}, December 2015, pp. 65--70.

\end{thebibliography}


\end{document}